# Belief Optimization for Binary Networks: A Stable Alternative to Loopy Belief Propagation

**Max Welling**
Gatsby Computational Neuroscience Unit
University College London
17 Queen Square
London, WC1N 3AR, U.K.

**Yee Whye Teh**
Department of Computer Science
University of Toronto
10 King's College Road
Toronto, M5S 3G5 Canada

## Abstract

We present a novel inference algorithm for arbitrary, binary, undirected graphs. Unlike loopy belief propagation, which iterates fixed point equations, we directly descend on the Bethe free energy. The algorithm consists of two phases, first we update the pairwise probabilities, given the marginal probabilities at each unit, using an analytic expression. Next, we update the marginal probabilities, by following the negative gradient of the Bethe free energy. Both steps are guaranteed to decrease the Bethe free energy, and since it is lower bounded, the algorithm is guaranteed to converge to a local minimum. We also show that the Bethe free energy is equal to the TAP free energy up to second order in the weights. In experiments we confirm that when belief propagation converges it usually finds identical solutions as our belief optimization method. The stable nature of belief optimization makes it ideally suited for learning graphical models from data.

## 1 INTRODUCTION

Belief propagation (BP) is an efficient local message passing protocol for exact inference on trees (Pearl, 1988). Applying the same rules to graphs with cycles, named loopy BP, has proven a successful strategy for approximate inference (Murphy et al., 1999). In particular, it was shown that the celebrated method of "turbo decoding" is equivalent to loopy BP on an appropriate graphical model (McEliece et al., 1998),(Frey and MacKay, 1997). Other applications can be found in image analysis (Freeman and Pasztor, 1998), (Frey, 1999).

An important drawback of loopy BP is that it can easily fail to converge (e.g. it may get stuck in limit cycles). Important progress in understanding the convergence properties and the quality of the approximation was made in (Weiss, 2000). But the most important breakthrough came with the observation that the fixed points of the loopy BP updates are exactly at the stationary points of the Bethe free energy (Yedidia et al., 2000), (Yedidia, 2000) (see also (Kabashima and Saad, 1998)). This did not only clarify the nature of the approximation made, it also opened up a large body of physics literature from which an interesting generalization could be derived (the generalized BP algorithm which minimizes the Kikuchi free energy).

An attractive feature of loopy BP is the fact that it performs local computations to find a solution which is approximately optimal globally. Solutions to the Bethe approximation can be understood as mean field solutions, in that pairs of nodes are interacting with a mean field generated from the rest of the system. Since this must be true for any pair of nodes, this leads to consistency equations equivalent to the BP updates. Yet another way of deriving the BP equations is through the "cavity method", where one computes the effect of taking one node and its connections out of the system (Opper and Winther, 2000).

The problem with a set of consistency equations is that they are not guaranteed to converge. A more stable approach is to minimize an objective function if there exists one. Since we know that BP converges to the stationary points of the Bethe free energy, it seems natural to derive an algorithm which minimizes it directly. This is precisely the approach taken in this paper. Progress can be made especially for binary networks, since they allow a convenient parameterization which avoids the usual need for Lagrange multipliers. Moreover, in the binary case we can find an analytic expression for the correlations $p(s_i = 1, s_j = 1)$ in terms of the marginals $p(s_i = 1)$ and $p(s_j = 1)$. This result can be used to formulate gradient descent or fixed point updates for the marginals in terms of the neighbouring marginals and the connecting weights.

In our framework, we are also able to show the equivalence between the TAP free energy and the Bethe free energy up to second order in the weights.

## 2　BELIEF OPTIMIZATION

In this section we will introduce the belief optimization (BO) algorithm. The model is represented by an undirected graphical model, where pairs of units are connected by weights $W_{ij}$. The units can take values $\{0,1\}$ and have biases $b_i$ (i.e. a Boltzmann machine). If a unit, say $v_j$, is observed ($v$ stands for "visible"), it will add an amount $W_{ij}v_j$ to the bias of a neighbouring unit $i$. We are interested in computing the marginal probability table $p_i$ of each hidden unit $h_i$, and the joint probability table $p_{ij}$ of each pair of neighbouring units $h_i$ and $h_j$.

We will use the following parameterization of the probability tables, which turns out to be convenient for binary variables,

$$p_{ij}(h_i=1, h_j=1) = \xi_{ij} \quad (1)$$
$$p_i(h_i=1) = q_i \quad (2)$$

All the other entries of the probability tables can be expressed in terms of this set of independent parameters,

$$p_{ij}(h_i=1, h_j=0) = q_i - \xi_{ij} \quad (3)$$
$$p_{ij}(h_i=0, h_j=1) = q_j - \xi_{ij} \quad (4)$$
$$p_{ij}(h_i=0, h_j=0) = \xi_{ij} + 1 - q_i - q_j \quad (5)$$
$$p_i(h_i=0) = 1 - q_i \quad (6)$$

It can also easily be checked that all marginalization constraints are satisfied, e.g.

$$\sum_{h_i=0,1} p_{ij}(h_i, h_j=1) = q_j \quad (7)$$
$$\sum_{h_i=0,1} p_{ij}(h_i, h_j=0) = 1 - q_j \quad (8)$$
$$\sum_{h_i,h_j=0,1} p_{ij}(h_i, h_j) = 1 \quad (9)$$

The main idea is now to write the Bethe free energy directly in terms of the above variables and minimize. Using the general expression for the Bethe free energy (see (Yedidia, 2000)) and the definitions above, we arrive at

$$F_b = E - S_1 - S_2 \quad (10)$$
$$E = -\sum_{(ij)} W_{ij}\xi_{ij} - \sum_i b_i q_i$$
$$-S_1 = \sum_i (1-z_i)[q_i \ln(q_i) + (1-q_i)\ln(1-q_i)]$$

$$-S_2 = \sum_{(ij)} \xi_{ij}\ln(\xi_{ij})$$
$$+ (\xi_{ij} + 1 - q_i - q_j)\ln(\xi_{ij} + 1 - q_i - q_j)$$
$$+ (q_i - \xi_{ij})\ln(q_i - \xi_{ij})$$
$$+ (q_j - \xi_{ij})\ln(q_j - \xi_{ij})$$

where $z_i$ denotes the number of neighbours of node $i$, and $(ij)$ denotes a link from node $i$ to node $j$.

We will first consider all the marginals $q_i$ fixed and equate the derivatives with respect to $\xi_{ij}$ to zero,

$$\frac{\partial F_b}{\partial \xi_{ij}} = -W_{ij} + \ln\left[\frac{\xi_{ij}(\xi_{ij} + 1 - q_i - q_j)}{(q_i - \xi_{ij})(q_j - \xi_{ij})}\right] = 0 \quad (11)$$

This can be rewritten as a simple quadratic equation,

$$\alpha_{ij}\xi_{ij}^2 - (1 + \alpha_{ij}q_i + \alpha_{ij}q_j)\xi_{ij} + (1+\alpha_{ij})q_iq_j \quad (12)$$

where we have defined,

$$\alpha_{ij} = e^{W_{ij}} - 1 \quad (13)$$

In addition to this equation we have to make sure that $\xi_{ij}$ satisfies the following bounds,

$$\max(0, q_i + q_j - 1) \leq \xi_{ij} \leq \min(q_i, q_j) \quad (14)$$

These bounds can be understood by noting that probabilities can not become negative. In appendix (A) we will prove the following lemma:

**Lemma 1** *There is exactly one solution to the quadratic equation (12) which satisfies the bounds (14). The analytic expression is given by,*

$$\xi_{ij} = \frac{1}{2\alpha_{ij}}\left(Q_{ij} - \sqrt{Q_{ij}^2 - 4\alpha_{ij}(1+\alpha_{ij})q_iq_j}\right)$$
$$Q_{ij} = 1 + \alpha_{ij}q_i + \alpha_{ij}q_j \quad (15)$$

*Moreover, $\xi_{ij}$ will never actually saturate one of the bounds.*

Note that for $\alpha_{ij} \to 0$ we have $\xi_{ij} = q_iq_j$ which is the correct limit[1]. This lemma is interesting since it allows one to estimate the correlations in a binary network given the marginals $q_i$.

In the other phase of the algorithm, we update the $q_i$, such that the free energy is guaranteed to decrease.

---

[1] For computational reasons it is sometimes convenient to use the following equivalent expression,

$$\xi_{ij} = \frac{1}{2}\left(R_{ij} - \text{sign}(\beta_{ij})\sqrt{R_{ij}^2 - 4(1+\beta_{ij})q_iq_j}\right)$$
$$R_{ij} = \beta_{ij} + q_i + q_j \quad \text{and} \quad \beta_{ij} = \frac{1}{\alpha_{ij}}$$

One way to achieve this, is to fix all $\xi_{ij}$ and all neighbouring $q_j$ of the marginal $q_i$ which is currently under consideration. It can be shown again that this is a convex optimization problem, with a (unique) solution inside the following bounds,

$$\max_{j \in N(i)} (\xi_{ij}) < q_i < \min_{j \in N(i)} (\xi_{ij} + 1 - q_j) \quad (16)$$

where $N(i)$ denotes the set of all neighbours of node $i$. However, there are regimes where the bounds become tight, and only little progress can be made at each step by alternating the $\xi$ and $q$ updates.

A more attractive procedure is to consider the $\xi_{ij}$'s as a function of the $q_i$'s and insert them back into the free energy, and then update the $q_i$'s. The advantage is that we do not need to consider the bounds (16), but simply have to make sure all $q_i$ lie between 0 and 1 (i.e. they define a probabilities). This can be achieved by reparameterizing,

$$q_i = \sigma(y_i) \quad (17)$$

where $\sigma$ stands for the sigmoid function and $y_i$ is unbounded. Taking derivatives with respect to $y_i$, we arrive at,

$$\frac{dF_b}{dy_i} = \left( \frac{\partial F_b}{\partial q_i} + \sum_{j \in N(i)} \frac{\partial F_b}{\partial \xi_{ij}} \frac{\partial \xi_{ij}}{\partial q_i} \right) q_i (1 - q_i) \quad (18)$$

$$\frac{\partial F_b}{\partial q_i} = -b_i + \ln \left[ \frac{(1-q_i)^{z_i - 1} \prod_{j \in N(i)} (q_i - \xi_{ij})}{q_i^{z_i - 1} \prod_{j \in N(i)} (\xi_{ij} + 1 - q_i - q_j)} \right]$$

and $\frac{\partial F_b}{\partial \xi_{ij}} = 0$ because $\xi_{ij}$ is at a minimum of $F_b$.

We can now use any gradient based optimization algorithm to minimize the free energy. Notice, that we can do gradient steps for all nodes $i$ simultaneously, instead of the sequential updates in the coordinate descent algorithm described before.

Alternatively, we can iterate the following set of fixed point equations for $q_i$,

$$q_i^* = \sigma \left( b_i + \ln \left[ \frac{q_i^{z_i} \prod_{j \in N(i)} (\xi_{ij} + 1 - q_i - q_j)}{(1 - q_i)^{z_i} \prod_{j \in N(i)} (q_i - \xi_{ij})} \right] \right) \quad (19)$$

As we will show in the next section, these fixed point equations exactly reduce to mean field updates if we retain terms linear in $W_{ij}$, and to TAP-updates if we retain terms quadratic in $W_{ij}$. They can therefore be understood a generalization of MF and TAP updates, which include higher order terms in $W_{ij}$. Unfortunately, fixed point equations are not guaranteed to converge and sometimes need a considerable amount of damping to avoid oscillations.

## 3 RELATION TO TAP AND THE SMALL WEIGHT EXPANSION

In this section we answer the question of whether there is a relationship between the Bethe free energy and the TAP free energy, given by,

$$Ftap = E - S1 - T \quad (20)$$

$$E = -\sum_{(ij)} W_{ij} q_i q_j - \sum_i b_i q_i \quad (21)$$

$$-S_1 = \sum_i [q_i \ln(q_i) + (1 - q_i) \ln(1 - q_i)]$$

$$-T = -\frac{1}{2} \sum_{(ij)} W_{ij}^2 \ q_i(1 - q_i) q_j (1 - q_j) \quad (22)$$

where $T$ is the TAP-correction to the mean field free energy.

Using the general expression $\frac{\partial F}{\partial W_{ij}} = -\mathbf{E}[h_i h_j]$, valid at the minimum of the free energy, the correlations become,

$$\xi_{ij}^{tap} = q_i q_j + W_{ij} \ q_i(1 - q_i) q_j (1 - q_j) \quad (23)$$

The fixed point equations can be derived by taking derivatives with respect to $q_i$,

$$\frac{\partial Ftap}{\partial q_i} = -\sum_{j \in N(i)} W_{ij} q_j - b_i + \log \left[ \frac{q_i}{1 - q_i} \right]$$

$$+ \frac{1 - 2q_i}{2} \sum_{j \in N(i)} W_{ij}^2 \ q_j (1 - q_j) \quad (24)$$

which can be used for gradient descent directly or from which the following fixed point equation can be derived,

$$q_i^* = \sigma \left( \sum_{j \in N(i)} W_{ij} q_j + b_i + \frac{1 - 2q_i}{2} \sum_{j \in N(i)} W_{ij}^2 \ q_j (1 - q_j) \right) \quad (25)$$

We will now formulate the following result,

**Lemma 2** *The TAP free energy is equal to the Bethe free energy up to order $O(W_{ij}^2)$ i.e.*

$$Fbethe = Ftap + O(W_{ij}^3). \quad (26)$$

The proof proceeds as follows. Since we have an analytic expression for the pairwise probabilities $\xi_{ij}^{bethe}$ as a function of the marginals $q_i$ and $q_j$ (see 15), we can simply insert that expression into the Bethe free energy and expand the result in powers of $W_{ij}$. After some algebra the above claim follows (see also (Bowman and Levin, 1982) and (Nakanishi, 1981) for similar results). Since we have shown that the free energies are equal up to second order in the weights, it is also clear that the fixed point equations (19) and (25) are equal up to second order, since they are derived from their respective free energies by taking derivatives with respect to $q_i$. By a similar argument we can show that the correlations (15) and (23) are equal up to first order in the weights. When only terms linear in the weights are retained, the Bethe free energy reduces to the mean field free energy.

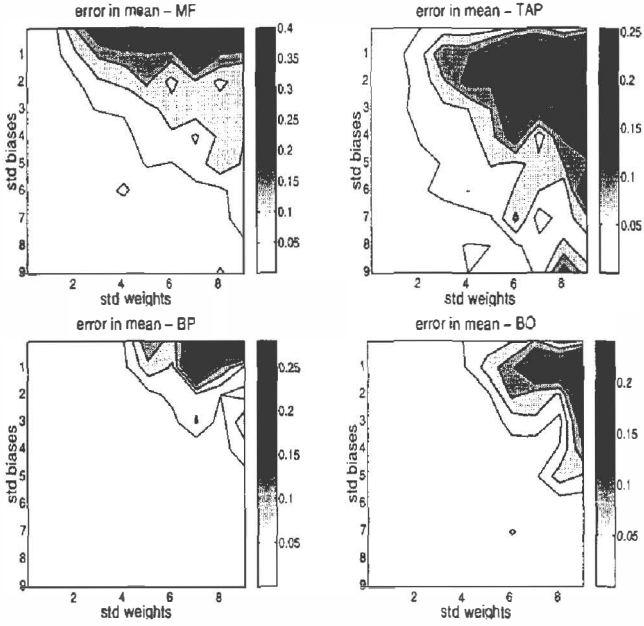

Figure 1: Absolute value of the difference between the exact means computed with the junction tree algorithm and MF, TAP, BP and BO respectively, averaged over all nodes. The network has 100 nodes placed on a square lattice (i.e. each node, except the boundary nodes has 4 neighbours). The scale of the weights varies over the horizontal axis from 0.1 to 10, sampled at intervals of 0.5. The scale of the biases varies similarly over the vertical axis.

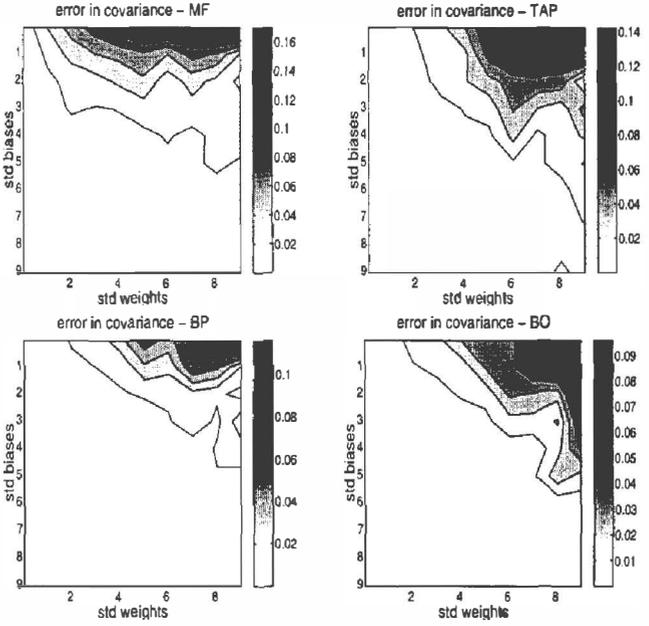

Figure 2: Absolute value of the difference between the exact covariances computed with the junction tree algorithm and MF, TAP, BP and BO respectively, averaged over all edges. Network specifics as in figure (1).

Since $\xi_{ij}$ is a function of $W_{ij}$ only, it is easy to see that the only terms which can contribute to the general expansion of $Fbethe$ are of the form $(W_{ij})^n$. Cross-terms, like $W_{ij}W_{jk}$, are not included. This result is consistent with the claim in (Yedidia, 2000) and (Georges and Yedidia, 1991) that the only terms contributing to the exact free energy of a binary undirected graph are "strongly irreducible". The latter means that if one draws a diagram for each term in the expansion, with a node for every index $i,j,k,...$, and a link between nodes $i$ and $j$ for every $W_{ij}$ in the term, then removing one node does not split the diagram in two. Since the Bethe free energy is exact on a tree, it must therefore sum up all strongly irreducible terms which connect two nodes together. The remaining ignored diagrams must then be all strongly irreducible diagrams which contain cycles[2].

## 4 EXPERIMENTS

To assess the quality of the approximation provided by BO, we compared the BO algorithm with 3 alternative inference methods, namely mean field (MF), TAP, and BP. The MF, TAP and BP fixed point equations were damped with a damping factor slowly increasing until 0.9. For BO we implemented an adaptive gradient descent algorithm. If no convergence was reached after 1000 updates for any of the methods, the program was halted. In the first experiment the units were placed on a $10 \times 10$ square grid for which the exact means and covariances $(\xi_{ij} - q_i q_j)$ can be computed using the junction tree algorithm. In a second experiment we used a $5 \times 5 \times 5$ cubic lattice with periodic boundary conditions, for which we used annealed Gibbs sampling (10000 samples) to compare our results against. The weights were drawn from a slightly super-Gaussian distribution to simulate the histogram one often observes when a Boltzmann machine has been trained on data. The biases were drawn from a Gaussian distribution and shifted by an amount $-\frac{1}{2}\sum_{j \in N(i)} W_{ij}$, such that in a network with no external evidence, a unit with zero (shifted) bias will have a mean of $\frac{1}{2}$. The scales of the weights and biases varied over a range from 0.1 to 10.

---

[2] We thank Jonathan Yedidia for pointing this out to us.

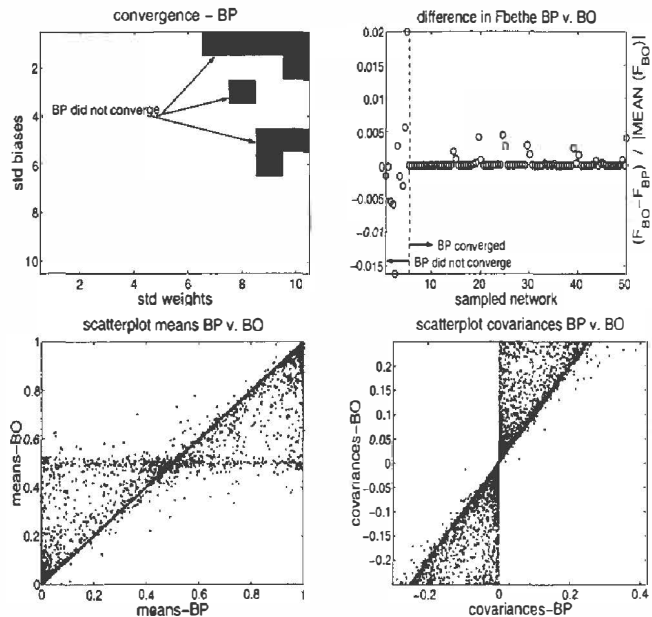

Figure 3: (top left) Convergence diagnosis of BP for the experiment on the 10 × 10 square lattice: black squares indicate failure to converge.(top right) Difference in the Bethe free energy between BO and BP. Cases for which BP did not converge are located on the left of the dashed line. (bottom) Scatterplots of the BP estimates (horizontal) and BO estimates (vertical) for the means (bottom left) and covariances (bottom right).

Figures (1) and (2) show the errors in the estimated means and covariances for the 4 different methods. It is evident that all methods become inaccurate when the weights are large and the biases are small. In this limit the strong correlations between the nodes are not modelled adequately by either of the methods since (1) MF does assume independence (2) TAP assumes small weights (3) the Bethe approximation assumes no cycles within the correlation distance (see the next section for a more detailed discussion). In figure (3) (top left) we show all instances where BP failed to converge, which are situated in the difficult regime mentioned above. The top right plot of figure (3) shows the differences between the Bethe free energies of BP versus BO. The bottom left and right figures show the scatterplots for the means and covariances respectively. Although most points populate the diagonal, these plots provide interesting information about the outliers which correspond to the difficult cases. While BP and BO usually agree on the sign of $q_i - \frac{1}{2}$, the BO estimates seem to be more biased towards $\frac{1}{2}$. Also in the case of the covariances BO and BP agree on the sign, but now the BP estimates are almost always smaller than the BO estimates in absolute value. Surprisingly, the relative free energies never differ by more than 0.02. For all converged cases (right of the dashed line) BP seems to have done slightly better in locating the minimum, but even for the non-convergent cases the differences are relatively small.

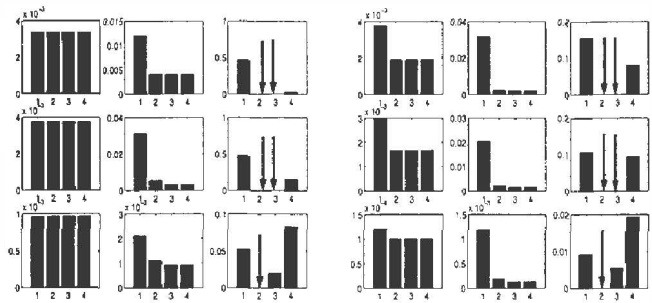

Figure 4: Absolute value of the difference between Gibbs sampling and 1 =MF, 2 =TAP, 3 =BP and 4=BO for the means (left) and covariances (right), averaged over all nodes and edges respectively. The network has 125 units placed on a cubic lattice with periodic boundary conditions (i.e. each node has 6 neighbours). The scale of the weights varies in the horizontal direction over $\{0.1, 1, 10\}$ (left to right). The size of the biases vary over the same values in the vertical direction (top to bottom). Arrows indicate that after 1000 updates there was no convergence. Notice that the smallest bar does not automatically correspond to the most accurate estimate, since results are compared with Gibbs sampling.

Figure (4) shows the results for the 5 × 5 × 5 lattice with periodic boundaries. The results are qualitatively similar, although the problem is more challenging due to the increased number of loops. Notice that in this case the smallest bar in figure (4) does not necessarily correspond to the best result, since Gibbs sampling is not necessarily the most accurate method.

From the above experiments we could not conclude a significant difference in performance between BP and BO. In the regime with large weights and small biases both methods fail, probably because the Bethe free energy is no longer an accurate approximation.

## 5  DISCUSSION

A notable difference between BP and BO is the fact that BP need not satisfy the marginalization constraints before it has converged. In contrast, BO is parameterized such that it will satisfy these constraints automatically. The above implies that the dynamics by which BP and BO try to minimize the Bethe free energy are of a very different nature. An undesirable property of BP, namely its failure to converge under certain circumstances, is certainly avoided by BO. However, the general conclusion from our experiments is that the Bethe approximation probably breaks down before any significant difference between the two methods shows up.

A situation where the freedom to violate the marginalization constraints before convergence may be crucial is when random variables are deterministically related. In this case probabilities turn into delta functions, energy barriers become infinitely high and the energy surface may become discontinuous. It is not hard to imagine how gradient desent could fail in these spaces. However, the success of turbodecoding proves that BP can effectively deal with these situations. More research is needed to find out whether direct minimization methods can be applied to this case as well.

The techniques discussed in this paper can also be applied to the Gaussian case. The details are worked out in appendix (B). For Gaussian BP (GaBP) it is important to notice that message updates do not necessarily maintain positive definiteness of the covariance matrix . This does not come as a surprise since it is a global constraint, while BP only performs local computations. As a consequence, the Bethe free energy is *not* always bounded from below and we have observed that exactly in these cases both GaBP and Gaussian BO (GaBO) do not converge. In all other cases GaBP and GaBO find the same answer experimentally. For a certain class of interactions (diagonally dominant) it was proved in (Weiss and Freeman, 1999) that GaBP always converges.

In this paper we have confirmed that both BP and BO perform poorly when the weights are large (strong correlations) but the biases remain small (little external evidence). One reason could be that the posterior energy surface has many modes in this regime, which may cause the lack of convergence for BP, and convergence to bad local minima for BO. It begs the question however whether the global minimum of Fbethe is just hard to find or whether the Bethe approximation itself breaks down. There are at least three regimes where the Bethe approximation should be accurate: for trees (exact), for small weights and for very large weights. For small weights, we can understand why loops do not have a significant influence, because they can at most contribute third order terms, e.g. $W_{ij}W_{jk}W_{ki}$. The larger the loop, the higher order the contribution and therefore the less effect it has on the final approximation. An alternative view on this is that evidence cycles around in the loops and is double counted as a consequence. When the weights are small, this information dies out before it has traversed a loop. In addition, external evidence tends to decrease dependencies between the units which will therefore soften the double counting effect and increase the accuracy of the Bethe approximation. As a rule of thumb one could therefore use that the Bethe approximation is probably reasonable when there are no loops with a circumference longer than the the correlation distance of the system. If this is not the case, one could use larger clusters within the Kikuchi approximation and the generalized BP algorithm (Yedidia et al., 2000) to improve performance. Finally, for very large weights (irrespective of the biases sizes), the energy term dominates the entropy term and the Bethe approximation should become exact. We have however not oberved good performance of either BP or BO in this regime, possible due to the many modes in the free energy surface.

It is our hope that the stable nature of BO will be usefull for learning graphical models from data, like the Boltzmann machine (BM). A major difficulty with learning is the fact that as learning progresses, the weights increase, and the usual approximations break down. Especially in the sleep phase of the BM, there is no evidence clamped on the nodes, and we enter the regime which was identified as the regime where *Fbethe* becomes a bad approximation. The introduction of the contrastive divergence learning objective (Hinton, 2000) has alleviated this problem at least partially, since the sleep phase is replaced by a phase where there is always a subset of the units clamped. In recent work, we have shown that the naive MF approximation works well in this case (Welling and Hinton, 2001). We are therefore eager to apply BO as an inference engine, inside the framework of contrastive divergence BM learning.

### Acknowledgements

We would like to thank Jonathan Yedidia, Manfred Opper, Yair Weiss, Zoubin Ghahramani and Geoffrey Hinton for interesting conversations, questions and suggestions.

## A PROOF OF LEMMA 1

We will first proof that there must be exactly one minimum inside the bounds (14) (i.e. not on the bounds).

First, we compute the second derivative with repect to $\xi_{ij}$,

$$\frac{\partial^2 F_b}{\partial \xi_{ij}} \quad (27)$$
$$= \frac{1}{\xi_{ij}} + \frac{1}{\xi_{ij} + 1 - q_i - q_j} + \frac{1}{q_i - \xi_{ij}} + \frac{1}{q_j - \xi_{ij}}$$
$$= \frac{1}{p_{ij}(1,1)} + \frac{1}{p_{ij}(0,0)} + \frac{1}{p_{ij}(1,0)} + \frac{1}{p_{ij}(0,1)} \geq 0$$

Also from $\frac{\partial F_b}{\partial \xi_{ij}}$ in (11) we see that at the lower boundary the derivative is $-\infty$ while at the upper boundary it is $+\infty$. Since the second derivative is always positive between the bounds and since the free energy is continuous between the bounds we infer that the free energy has exactly one minimum inside the bounds.

Next we proof that the positive root,

$$\zeta_{ij} = \frac{1}{2\alpha_{ij}}\left(Q_{ij} + \sqrt{Q_{ij}^2 - 4\alpha_{ij}(1+\alpha_{ij})q_iq_j}\right)$$
$$Q_{ij} = 1 + \alpha_{ij}q_i + \alpha_{ij}q_j \qquad (28)$$

to the quadratic equation (12) is always located outside the bounds (except for $\alpha_{ij} = 0$ when the equation is degenerate). We can assume without loss of generality that $q_i \geq q_j$.

For $\alpha_{ij} = 0$ we have that the quadratic equation reduces to,

$$-\xi_{ij} + q_iq_j = 0 \qquad (29)$$

with the obvious solution located between the bounds (14). For $\alpha_{ij} > 0$ we will use the fact that,

$$Q_{ij}^2 - 4\alpha_{ij}(1+\alpha_{ij})q_iq_j$$
$$= 1 + 2\alpha_{ij}q_i(1-q_j) + 2\alpha_{ij}q_j(1-q_i) + \alpha_{ij}^2(q_i - q_j)^2$$
$$\geq 1 + 2\alpha_{ij}[q_i(1-q_j) + q_j(1-q_i)]$$
$$\geq 0$$

(this result is actually valid for all possible $\alpha_{ij}$, i.e. in the range $(-1,\infty)$) The above result can now be used to prove,

$$\zeta_{ij} \geq \frac{1}{2\alpha_{ij}}(1 + \alpha_{ij}q_i + \alpha_{ij}q_j)$$
$$\geq \frac{1}{2\alpha_{ij}} + q_j$$
$$\geq q_j \qquad (30)$$

which is always larger than the upper bound. Finally, for $\alpha_{ij} \in (-1, 0)$, we will use the fact that,

$$Q_{ij}^2 - 4\alpha_{ij}(1+\alpha_{ij})q_iq_j \geq Q_{ij} \qquad (31)$$

with $Q_{ij}$ defined in (28). This can be used to prove,

$$\zeta_{ij} \leq \frac{1}{\alpha_{ij}} + q_i + q_j$$
$$\leq -1 + q_i + q_j$$

which is always smaller than the lower bound.

Therefore, since we know one of the solutions must be located at the minimum between the boundaries, and the positive root is always located outside the boundaries, we have proven that the negative root is precisely the valid solution, located at the minimum of the free energy, inside the boundaries.

## B GAUSSIAN BELIEF OPTIMIZATION

Let $\mu_i$ denote the mean of node $i$, $V_i$ denote the variance of node $i$, $V_{ij}$ denote the covariance between node $i$ and node $j$, $b_i$ the bias at node $i$ and $W_{ij}$ the interaction strength between node $i$ and node $j$. Up to constant terms, the Bethe free energy is given by,

$$F_b = E - S_1 - S_2 \qquad (32)$$
$$E = \sum_{(ij)} W_{ij}(V_{ij} + \mu_i\mu_j)$$
$$+ \frac{1}{2}\sum_i W_{ii}(V_i + \mu_i^2) + b_i\mu_i$$
$$S_1 = \sum_i \frac{1}{2}(1 - z_i)\log(V_i)$$
$$S_2 = \sum_{(ij)} \frac{1}{2}\log(V_iV_j - V_{ij}^2)$$

The problem of solving for the means $\mu_i$ decouples from the problem of solving for the covariance-matrix. The derivatives are given by,

$$\frac{\partial F_b}{\partial \mu_i} = \sum_{j \in N(i)} W_{ij}\mu_j + W_{ii}\mu_i + b_i \qquad (33)$$

which can be used in a simple gradient descent algorithm to solve for the $\mu_i$. Note that since the entropy, which is the only approximate term in the free energy, is independent of the $\mu_i$, and since the covariances are decoupled from the means in the energy, it follows immediately that the means are exact at the minimum of the free energy.

Taking derivatives with respect to the covariances $V_{ij}$,

$$\frac{\partial F_b}{\partial V_{ij}} = W_{ij} + \frac{V_{ij}}{(V_iV_j - V_{ij}^2)} \qquad (34)$$

we find the following quadratic equation,

$$W_{ij}V_{ij}^2 - V_{ij} - W_{ij}V_iV_j = 0 \qquad (35)$$

This quadratic equation has exactly one solution in the allowed region,

$$-\sqrt{V_iV_j} \leq V_{ij} \leq \sqrt{V_iV_j} \qquad (36)$$

namely,

$$V_{ij} = \frac{1}{2W_{ij}} - \text{sign}(W_{ij})\sqrt{\left(\frac{1}{2W_{ij}}\right)^2 + V_iV_j} \qquad (37)$$

The proof that this is the only viable solution proceeds as in appendix A. We will now eliminate $V_{ij}$ from the free energy and insert $V_{ij}$ as a function of $V_i$ into the free energy and take derivatives with respect to $V_i$. Since $V_i$ has to remain positive we reparameterize,

$$V_i = e^{y_i} \qquad (38)$$

and take derivatives with respect to $y_i$,

$$\frac{dF_b}{dy_i} = \left( \frac{\partial F_b}{\partial V_i} + \sum_{j \in N(i)} \frac{\partial F_b}{\partial V_{ij}} \frac{\partial V_{ij}}{\partial V_i} \right) V_i \qquad (39)$$

$$\frac{\partial F_b}{\partial V_i} = \frac{1}{2} \left( W_{ii} + \frac{z_i - 1}{V_i} - \sum_{j \in N(i)} \frac{V_j}{(V_i V_j - V_{ij}^2)} \right)$$

and $\frac{\partial F_b}{\partial V_{ij}} = 0$ since $V_{ij}$ is at a minimum of the Bethe free energy. Any gradient based minimization algorithm can then be used to solve for $V_i$.

We have found that this set of update equations will give exactly the same solution as the Gaussian BP (GaBP) fixed point iterations, derived in (Weiss and Freeman, 1999), provided both algorithms converge. Surprisingly, even Gaussian BO (GaBO) does not always converge, due to the fact that the Bethe free energy is not always bounded from below. This can be understood by observing that the term $\sum_{(ij)} W_{ij} V_{ij}$ can become arbitrary negative, while under certain conditions the term $\frac{1}{2} \sum_i W_{ii} V_i$ cannot make up for this. The problem is that both GaBP and GaBO cannot impose the positive definiteness constraint on the covariance matrix in general, since this is a global constraint, while we are performing only local computations. From emperical studies we have found that GaBP and GaBO either converge to the same answer, or both fail to converge. This leads us to conjecture the following,

**Conjecture 1** *Gaussian Belief Propagation converges if and only if the free energy is bounded from below.*

It was shown in (Weiss and Freeman, 1999) that for diagonally dominant weight matrices the free energy is always bounded from below and GaBP always converges.